\pdfoutput=1

\documentclass[11pt]{article}

\usepackage{ACL2023}

\usepackage{times}
\usepackage{latexsym}

\usepackage[T1]{fontenc}

\usepackage[utf8]{inputenc}
\usepackage[symbol]{footmisc}
\usepackage{microtype}

\usepackage{inconsolata}
\usepackage{makecell}
\usepackage{times}
\usepackage{latexsym}
\usepackage{multirow}
\usepackage{booktabs}
\usepackage{multicol}
\usepackage{pdfpages}
\usepackage{xspace}
\usepackage{subfigure}
\usepackage[T1]{fontenc}
\usepackage[font=small,labelfont=bf,tableposition=top]{caption}
\usepackage[T1]{fontenc}
\usepackage{amssymb}
\usepackage{booktabs,makecell}
\usepackage[utf8]{inputenc}

\usepackage{microtype}
\usepackage{lipsum}

\usepackage{inconsolata}
\usepackage{amsmath}
\usepackage{pifont}
\usepackage{comment}
\definecolor{mypink}{rgb}{0.858, 0.188, 0.478}

\definecolor{kmy-color}{rgb}{0.28, 0.58, 0.28}

\newcommand{\tabincell}[2]{\begin{tabular}{@{}#1@{}}#2\end{tabular}}
\newcommand{\dataset}{\textsc{FollowupQG}\xspace}

\usepackage{array}
\newcommand{\PreserveBackslash}[1]{\let\temp=\\#1\let\\=\temp}
\newcolumntype{C}[1]{>{\PreserveBackslash\centering}p{#1}}
\newcolumntype{R}[1]{>{\PreserveBackslash\raggedleft}p{#1}}
\newcolumntype{L}[1]{>{\PreserveBackslash\raggedright}p{#1}}

\title{\dataset: Towards Information-Seeking Follow-up \\ Question Generation}

 \author{ Yan Meng$^{*}$ \\ University of Amsterdam  \\  \texttt{y.meng@uva.nl}
         \And  
         Liangming Pan$^{*}$\\ University of California, Santa Barbara \\ \texttt{liangmingpan@ucsb.edu}
         \AND
         Yixin Cao \\ Singapore Management University \\ \texttt{caoyixin2011@gmail.com} 
         \And      Min-Yen Kan \\ National University of Singapore \\ \texttt{kanmy@comp.nus.edu.sg}       }

\begin{document}
\maketitle

\def\thefootnote{$\ast$}\footnotetext{Authors contributed equally.}
\def\thefootnote{\arabic{footnote}}

\begin{abstract}
Humans ask follow-up questions driven by curiosity, which reflects a creative human cognitive process. We introduce the task of \textit{real-world information-seeking follow-up question generation (FQG)}, which aims to generate follow-up questions seeking a more in-depth understanding of an initial question and answer. We construct~\dataset, a dataset\footnote{Data available at \url{https://github.com/vivian-my/FollowupQG}} of over 3K real-world (initial question, answer, follow-up question) tuples collected from a Reddit forum providing layman-friendly explanations for open-ended questions. 

In contrast to existing datasets, questions in~\dataset use more diverse pragmatic strategies to seek information, and they also show higher-order cognitive skills (such as \textit{applying} and \textit{relating}). We evaluate current question generation models on their efficacy for generating follow-up questions, exploring how to generate specific types of follow-up questions based on \textit{step-by-step} demonstrations. Our results validate~\dataset as a challenging benchmark, as model-generated questions are adequate but far from human-raised questions in terms of informativeness and complexity.

\end{abstract}

\section{Introduction}

Question asking is considered a fundamental cognitive process. People typically ask concise and natural questions to seek information~\cite{ram1991theory}. \textit{Question Generation} (QG) has recently gained much interest, targeting the study of how intelligent systems can generate relevant questions. This can evaluate the cognitive reasoning ability of models while benefiting many downstream tasks, such as generating assessments for course materials in education~\cite{DBLP:journals/corr/abs-2205-01730} and enriching training data for question answering~\cite{DBLP:conf/naacl/PanCXKW21}. 

\begin{figure}[!t]
\centering
\subfigure[\textbf{Answer-aware} Question Generation]
{
	\begin{minipage}[t]{\linewidth}
	\centering
	\includegraphics[width=6.5cm]{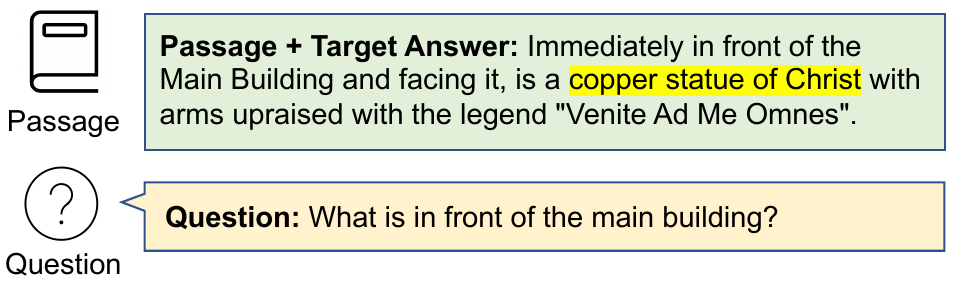}
	\end{minipage}
	\label{fig:answer-aware-example}
}
\subfigure[\textbf{Information-seeking} Question Generation]
{
	\begin{minipage}[t]{\linewidth}
	\centering
	\includegraphics[width=6.5cm]{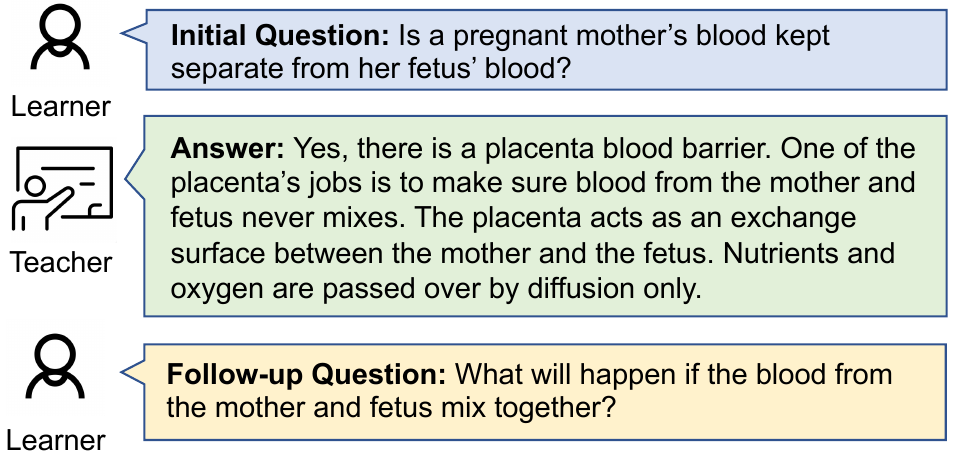}
	\end{minipage}
	\label{fig:information-seeking-example}
}
\vspace{-0.2cm}
\caption{Examples of (a) \textit{answer-aware QG} and (b) \textit{information-seeking QG}.}
\vspace{-0.6cm}
\label{fig:dataset_example_1}
\end{figure}

Existing works~\cite{DBLP:conf/emnlp/DuanTCZ17,DBLP:conf/emnlp/ZhaoNDK18,DBLP:conf/acl/PanXFCK20,DBLP:conf/acl/GhanemCDOF22} focus on generating simple factoid questions, while few works to date target complex practical questions. The task of QG is often framed as generating questions from a source text and a specific target answer from reading comprehension datasets like SQuAD~\cite{DBLP:conf/emnlp/RajpurkarZLL16}, as exemplified by Figure~\ref{fig:answer-aware-example}. Although useful in practical applications, such generated questions are quite different from actual human questions. First, they do not reflect the information-seeking nature of human question-asking, since the model already knows the answer beforehand. Second, they also do not reflect the creative human cognitive process in question-asking such as inferences and synthesis.

To bridge this gap, we propose the task of \textit{real-world information-seeking follow-up question generation (FQG)}, which aims to generate \textit{follow-up questions} that seek new information given the \textit{initial question} and the \textit{human-provided answer}. For example, the follow-up question in Figure~\ref{fig:dataset_example_1} extends the provided answer to a reasonable counterfactual situation. Conventional follow-up question generation works focus on benefiting multi-hop reasoning QA systems~\citep{Malon2020GeneratingFQ} or generating multi-turn conversational questions~\citep{reddy-etal-2019-coqa,richardson-etal-2023-learning}. In contrast, our task is more practical and challenging, since it requires a higher level of cognition to know what one does not know~\citep{miyake1979ask}. First, it demands a deep comprehension of the teacher-provided answer, identifying the uncertainty or gaps in knowledge; and second, applying high cognitive skills such as analogy to generate a meaningful follow-up question.


In this paper, we construct a dataset, \dataset, containing 3,790 real-world (initial question, answer, follow-up question) tuples. We collect the data from the Reddit forum \textit{Explain Like I'm Five}\footnote{\url{https://www.reddit.com/r/explainlikeimfive/}} which contains real-life questions and self-contained answers. The layperson-friendly nature of this forum makes the question and answer highly comprehensible, serving as a suitable context for follow-up question generation. We further ask crowd-workers to select relevant follow-up questions from replies to the answer as they are real curiosity-driven questions by humans. Our data analysis shows that~\dataset captures a variety of high cognitive skills in question-asking, such as relating and causal inference.

We establish benchmarks on this data using GPT-Neo~\citep{gpt-neo}, BART~\citep{DBLP:conf/acl/LewisLGGMLSZ20} and T5~\citep{DBLP:journals/jmlr/RaffelSRLNMZLL20}. Automatic and human evaluation reveals that the best model can generate fluent follow-up questions. However, they still fall short of human questions in terms of semantic validity, complexity, and informativeness. Also, we find that $\sim$30\% of the generated questions do not seek new information. 

We note that one limitation of fine-tuning pre-trained language models on QG task is to control \textit{what to ask} and \textit{how-to ask}. Inspired by recent prompting methods on large language models~\cite{Wei2022ChainOT, saha2023summarization},  we investigate on \textit{chain-of-thought} prompt-based learning via GPT family\footnote{ChatGPT, GPT-3.5, GPT-4}, and observe that incorporating an intermediate reasoning chain can better control the models to ask specific types of questions compared to the standard promoting. However, there is still a large improvement in generating specific high-level questions. These observations make \dataset a challenging benchmark for advancing QG.



\section{Related Work}
Question Generation (QG) aims to automatically generate questions from textual input. Existing QG studies~\cite{DBLP:conf/acl/DuSC17,DBLP:conf/acl/CardieD18,DBLP:conf/emnlp/NemaMKSR19,DBLP:conf/acl/PanXFCK20,DBLP:journals/corr/abs-2110-08175} are typically trained and evaluated on reading comprehension benchmarks such as SQuAD~\cite{DBLP:conf/emnlp/RajpurkarZLL16} and HotpotQA~\cite{DBLP:conf/emnlp/Yang0ZBCSM18}. 
Questions in those datasets are designed to test machine's reading comprehension ability, which fail to reflect the information-seeking nature of human question-asking. This gap has led to work on ``answer-agnostic'' question generation~\cite{DBLP:conf/acl/SubramanianWYZT18,DBLP:conf/aaai/WangWFLH19,pan2019recent}, in which the target answer is not given to the model as input. However, the sources of data are still reading comprehension datasets and the generated questions are still required to be answerable by the input source, which are quite different with the follow-up questions in our work, which aims to seek for unknown information from known knowledge. 
 
To explore the generation of real human-like information-seeking questions, prior works have investigated generating clarification questions for forum posts in \textit{StackExchange}~\cite{DBLP:conf/acl/DaumeR18,DBLP:conf/acl/KumarB20}, Amazon product reviews~\cite{DBLP:conf/naacl/RaoD19,DBLP:conf/naacl/MajumderRGM21}, and online courses~\cite{DBLP:conf/icwsm/ChenYHH18}. However, clarification is only one of the pragmatic goals in asking follow-up questions. \dataset covers broader types of information-seeking behaviors beyond clarification, such as association, analogy, critical evaluation, and generalization. In addition, instead of focusing on restricted and highly-technical domains like StackExchange and Amazon products, we select \textit{Explain Like I'm Five} as the underlying data source, which covers a boarder range of real-life topics~\cite{DBLP:conf/acl/FanJPGWA19}. 


The closest prior work is InquisitiveQG~\cite{DBLP:conf/emnlp/KoCHDL20}. They asked crowd-workers to write follow-up questions for news articles and trained models for follow-up question generation. However, our analysis reveals that  crowd-sourced questions in InquisitiveQG are typically shallow in reasoning and biased towards monotonous cognitive skills, in contrast with our natural follow-up questions collected from the web. In addition, our work focuses on a scenario different with InquisitiveQG but common in real-life: asking follow-up questions based on the initial question and its answer. 

\begin{figure}[!t]
	\centering
	\includegraphics[width=7.5cm]{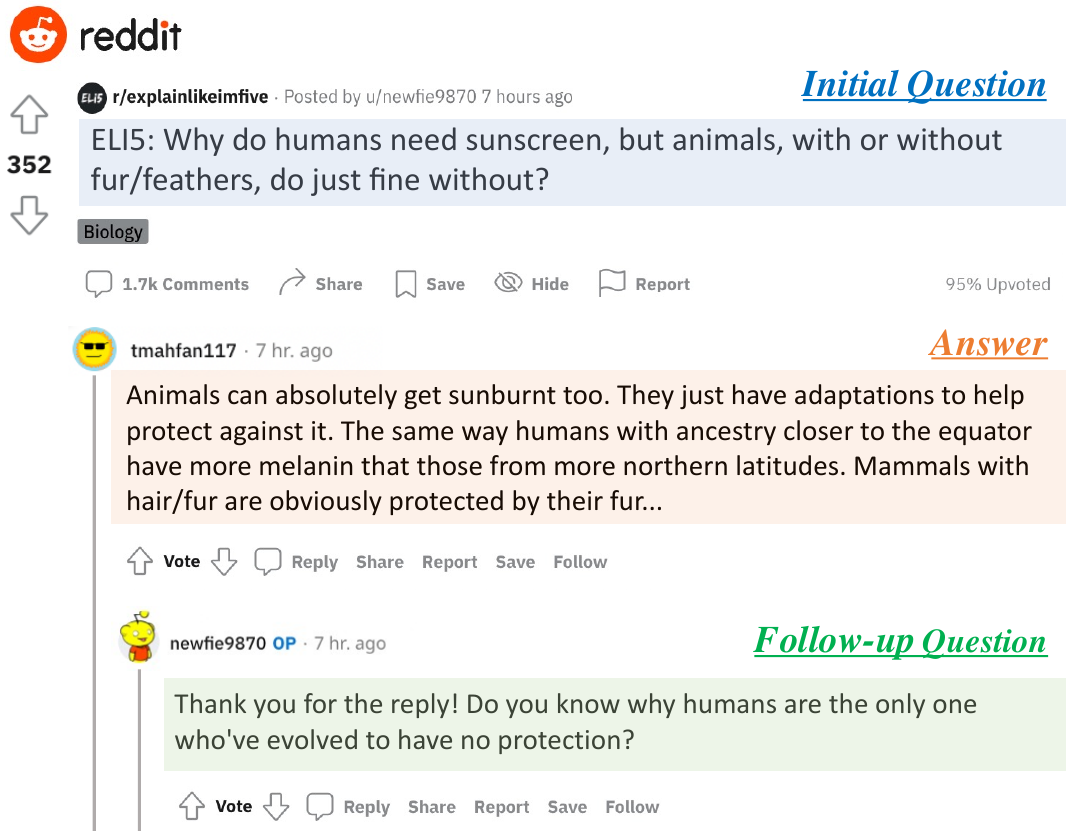}
    \caption{Sample \textit{Explain like I'm five} (ELI5) forum.}
    \label{fig:dataset_example_2}
    \vspace{-0.2cm}
\end{figure}

\section{The \dataset Dataset}

We construct the \dataset dataset as follows. The follow-up questions and the source documents are collected from \textit{Reddit} \footnote{License of usage: \url{https://www.redditinc.com/policies/data-api-terms} }(\S~\ref{ss:3.1}). We first collect around 200,000 posts that contains question, answer, and replies to the answer with a site-specific web crawler. Then, we automatically select data samples that contain follow-up questions in \S~\ref{ss:3.2}. Afterward, the selected 10,890 data samples are further validated by online workers from Amazon Mechanical Turk (AMT) (\S~\ref{ss:3.3}). The final dataset contains 3,790 high-quality samples. 

\subsection{Data Sources} \label{ss:3.1}

To gather real-world information-seeking questions, we initially explored several websites which provide forums to ask open-ended questions, such as Quora, Khan Academy, as well as numerous Reddit forums (subreddits). After a careful comparison, we choose to focus on the subreddit \textit{Explain Like I'm Five} (ELI5), where users are encouraged to provide answers which are comprehensible by a five year-old child. ELI5 is appealing because the questions are close to real-life and the answers are self-contained, and thus rely less on prior specialized knowledge. 
Their high comprehensibility makes the question and answer suitable to serve as the context of follow-up question generation.

\subsection{Data Collection} \label{ss:3.2}
A thread in the ELI5 forum (Figure~\ref{fig:dataset_example_2}) usually consists of: (1) a thread title, usually in question format and is considered as the \textit{Initial Question}, (2) a vote that measures the quality of the thread, (3) top-level comments, most of which are detailed answers to the initial question, and (4) replies to the top-level comments, and many of them are asking follow-up questions to the answer. 

\citet{DBLP:conf/acl/FanJPGWA19} have collected a large amount of (question, answer) pairs from ELI5 for question-answering. However, we could not reuse their corpus since they did not collect the follow-up questions. Therefore, we implement a site-specific web crawler to massively crawl data from the ELI5 forum. The crawler is built on \textit{Pushshift API} and \textit{Reddit API}, which give access to the post ID, body, vote, and the comments. We restrict the data collection size to 200,000 and only collect the first three levels of comments. 

We then define rules based on regular expressions to automatically filter out the invalid samples in the crawled data. A thread is considered invalid if: 1) its thread title is not a question, 2) the answer is not self-contained (shorter than 30 characters\footnote{A pilot study is conducted to check the answers ranging from 10 to 50 characters, and results show that answers shorter than 30 characters are generally less informative.}) or receive low votes, or 3) the replies to the answer do not contain any question. After applying this automatic filtration, 10,890 data samples remain.

\begin{table*}[!t]
    \centering
    \resizebox{\textwidth}{!}{%
    \begin{tabular}{p{160pt}p{210pt}C{50pt}C{20pt}}
    \hline
        \small \textbf{Context} &  \small \textbf{\dataset Examples}&  \small \textbf{Category} & \small \textbf{Ratio}  \\
        \hline
        \multirow{5}{=}{\small \textbf{Initial question:} Is a pregnant  mother's blood \textcolor{blue}{kept separate} from her fetus' blood? \\ 
        \small \textbf{Answer:} Yes, there is a \textcolor{purple}{placenta} blood barrier. One of the placenta's jobs is to make sure blood from the mother and fetus never \textcolor{red}{mixes}. The placenta acts as an \textcolor{cyan}{exchange surface} between the mother and the fetus. Nutrients and oxygen are passed over by \textcolor{orange}{diffusion} only.} & \small What does the placenta \textcolor{cyan}{exchange}?&\small Definition & \small 23\% \\
        
         &\small Why is it that nutrients and oxygen can only be passed over by \textcolor{orange}{diffusion}?& \small Interpretation & \small 38\%\\
      
          &\small What will happen if the blood \textcolor{red}{mixes}? & \small Counterfactual & \small 19\%\\

          &\small Will the \textcolor{purple}{placenta} still function if the woman is not pregnant?   & \small Relating &\small 6\%\\
         
          &\small Could someone give some suggestions on keeping the blood completely \textcolor{blue}{separate}?  & \small  Creative & \small 11\% \\
          & & & \\
         \hline
    \end{tabular}
    }
    \vspace{0.1cm}
    \caption{Question examples of different types of pragmatic functions in \dataset. The question-triggering text spans in the context are highlighted in different colors. }
    \vspace{-0.8cm}
    \label{tab:qe}
\end{table*}

\subsection{Crowd-Sourced Data Validation} \label{ss:3.3}

We find that the automatic-filtered data samples are still noisy. Especially, some replies in question-format are irrelevant to the initial question and answer or contain toxic or offensive contents.
To ensure that our final corpus contain high quality follow-up questions, we design a crowd-sourcing task for data validation. We release 10,000 HITs (Human Intelligence Tasks) on the AMT platform, evenly divided into 10 batches. Each HIT presents the crowd-worker with one data sample of \textit{(initial question, answer, follow-up question)}. To conduct human validation, we ask workers to answer three questions as follows: 


\vspace{0.15cm}
\noindent $\bullet$ \textbf{Q1: }Is the follow-up question a complete question asking for new information?

\noindent $\bullet$ \textbf{Q2: }Does the data sample contain controversial topics, such as racism, hate speech, sexual topics, or offensive comments? 

\noindent $\bullet$ \textbf{Q3: }What is the relatedness of the follow-up question to the initial question and the answer?, where workers use the 5-point classification set of ``strongly related'', ``related'', ``slightly related'', or ``not related''. 
\vspace{0.15cm}

\noindent To select qualified workers, we restrict our task to workers who are located in five native English-speaking countries\footnote{Australia, Canada, Ireland, United Kingdom, USA.} and who maintain an approval rating of at least 90\%. To ensure the annotations fulfil our guidelines, we give ample examples in our annotation interface with detailed explanations to help workers understand the requirements. The detailed annotation guidelines are in Appendix \ref{sec:appendix}. Each data sample is annotated by two different workers. We find substantial agreement between annotators, with an average Cohen's Kappa is 0.78, where the inter-annotator Kappa for Q1, Q2, and Q3 are 0.80, 0.61 and 0.92, respectively. 

To evaluate the quality of annotation, we add 50 test samples to each batch of HITs. We get an average test accuracy of 0.73 for all 10 batches, indicating the high-quality of the data annotation. In the end, 112 workers participated in the task, with 96.35\% average acceptance rate. The average completion time for one HIT is around 40 seconds, and we set payment at USD 1.00/HIT. To construct the final dataset, we retain only the samples that are annotated as high-quality\footnote{Choosing ``Yes'' answer for Q1, ``No'' answer for Q2, and choosing ``strongly related'' or ``related'' for Q3.} by both annotators, resulting in 3,790 instances. We randomly select 2,790 for training, 500 for validation, and 500 for testing.

\section{Data Analysis}
The pragmatic functions of human-raised follow-up questions and their required cognitive skill levels are crucial for understanding the mechanism of human question-asking. These factors should be studied for building efficient question generators. In \S~\ref{ss:4.1}, we first characterize the pragmatic functions of questions in \dataset in accordance with the cognitive skills defined Bloom's Revised Taxonomy~\cite{anderson2001taxonomy}. Then in \S~\ref{ss:4.2}, by comparing against existing datasets, we will show that the questions in \dataset are of higher level and have richer pragmatic functions. 

\begin{table*}[!t]
    \centering
    \resizebox{\textwidth}{!}{%
    \begin{tabular}{L{4cm}C{0.5cm}C{0.5cm}C{0.5cm}C{0.5cm}C{0.5cm}C{0.5cm}C{0.5cm}C{1cm}C{1cm}C{1cm}}
    \toprule
        \small{\multirow{2}{*}{Dataset}} & \multicolumn{2}{c}{\small{Avg. of Words}} & \multicolumn{5}{c}{\small{Distribution of Cognitive Skills} }  & \multicolumn{3}{c}{\small Most Frequent Question Types }  \\
        & \small{ Ques.} & \small Doc. & \small {Rem.} & \small{Und.} & \small{App.} & \small{Anal.} & \small{Crea.} & \small 1st & \small 2nd & \small 3rd  \\ 
        \cmidrule(lr){1-1} \cmidrule(lr){2-3} \cmidrule(lr){4-8} \cmidrule(lr){9-11}
        \small\textbf{\dataset} & \small\textbf{43.6} & \small 143.5 &  \small 23 & \small 38 & \small\textbf{19} & \small 6 & \small\textbf{11} & \small other & \small why & \small how  \\
        \hline 
        \small SQuAD \citep{DBLP:conf/acl/RajpurkarJL18}   & \small 9.9 & \small 134.8 & \small \textbf{100} & \small 0 & \small 0 & \small 0 & \small 0 & \small what & \small how & \small when \\ 
        \hline
        \small LearningQ \citep{DBLP:conf/icwsm/ChenYHH18} & \small 16.9 & \small \textbf{1729.5} & \small 18 & \small\textbf{56} & \small 13 & \small\textbf{15} & \small 3 & \small why & \small other & \small what \\
        \hline
         \small InquisitiveQG \citep{DBLP:conf/emnlp/KoCHDL20} & \small 7.1 & \small 150.4 & \small 46 & \small 49 & \small 5 & \small 0 & \small 0 & \small what & \small why & \small how  \\
        \bottomrule
    \end{tabular}}
    \vspace{0.2cm}
    \caption{Descriptive features and statistics of \dataset and the datasets in comparison. We follow the Bloom's Taxonomy~\cite{anderson2001taxonomy} to define the cognitive skills of questions. \textbf{Rem.}: Remembering; \textbf{Und.}: Understanding; \textbf{App.}: Applying; \textbf{Anal.}: Analyzing; \textbf{Crea.}: Creating. For question types, we follow~\citet{DBLP:conf/acl/Liu} to categorize questions based on the interrogative word and define 9 question types: who, where, when, why, which, what, how, boolean, other.}
    \vspace{-0.3cm}
    \label{tab:compare}
\end{table*}

\subsection{Categories of follow-up questions} \label{ss:4.1}

We analyze 800 questions randomly sampled from our dataset and find that most follow-up questions fall into one of the following five categories that correspond to different cognitive levels in Bloom's Taxonomy. We show examples of each category in Table~\ref{tab:qe}, where question-triggering text spans in the context are highlighted. 

\vspace{0.2cm}
\noindent $\bullet$ \textbf{Definition}: 23.6\% of questions seek clarifications for the definition or meaning of entities or facts in the context. Examples are: 
\textit{What is the definition of ...?} We map these to the \textit{Remembering} level in Bloom's taxonomy.

\vspace{0.2cm}
\noindent$\bullet$ \textbf{Interpretation}: 38.9\% of questions seek interpretations for reasons, means, goals, or background information to gain a deeper understanding of the answer. Examples are: \textit{Could you explain the reason ... ?}
They correspond to the \textit{Understanding} level in Bloom's taxonomy.

\vspace{0.2cm}
\noindent$\bullet$ \textbf{Counterfactual}: 18.7\% of questions apply the learned knowledge in the answer to a reasonable counterfactual case. Examples are: \textit{What will happen if ... ?} These mostly correspond to the \textit{Applying} level in Bloom's taxonomy. 

\vspace{0.2cm}
\noindent$\bullet$ \textbf{Relating}: 6.3\% of questions ask patterns or relationships between existing examples in the context and other related cases, which belong to \textit{Analysis} level in Bloom's taxonomy. Examples are like: \textit{What is the relationship between ... ?}

\vspace{0.2cm}
\noindent$\bullet$ \textbf{Creative}: 11.1\% of questions require the asker's creative thinking to invent new solutions or suggestions for learned facts in the context. They belong to \textit{Creating} level in Bloom's taxonomy. Examples are: \textit{Could ... be changed to improve ...?}

\vspace{0.2cm}
\noindent$\bullet$ \textbf{Others}: 1.3\% are rhetorical questions, \textit{e.g.}, expressing surprise by asking \textit{Oh, really?}.
\vspace{0.2cm}

In summary, 62.5\% of human-raised follow-up questions are clarification questions asking for definitions and interpretation, while 36.1\% of questions require higher-level cognitive thinking. This shows that \dataset has a relatively high proportion of questions that promote deep reasoning, considering the fact that asking deep questions is challenging for humans, revealed by prior studies~\cite{graesser1994question,dillon1988questioning}. 

\begin{table*}[!t]
    \centering
   \resizebox{0.9\textwidth}{!}{%
    \begin{tabular}{p{2.4cm}|cccccc|cccc}
         \toprule
         \small\textbf{Models} & \small\textbf{B1.}  & \small\textbf{B2.} & \small\textbf{B3.} & \small\textbf{B4.} & \small\textbf{MET.} &\small\textbf{ROU}$_L.$ & \small\textbf{ FLU.} & \small \textbf{REL.} & \small \textbf{COM.} & \small \textbf{INF.}\\
          \hline
         \small BART &  \small \textbf{17.22} & \small \textbf{7.11} & \small \textbf{3.89} & \small \textbf{2.61} & \small \textbf{8.00} & \small \textbf{13.35} & \small 4.54 & \small \textbf{0.99} & \small 1.36 & \small \textbf{1.31} \\
        \small T5 & \small 13.69 & \small 4.32 & \small 1.85 & \small 1.02 & \small 5.79 & \small 12.49 & \small \textbf{4.89} & \small 0.95 & \small \textbf{1.51} & \small 1.26 \\
        \small GPT-Neo & \small 14.08 & \small 4.09 & \small 1.89 & \small 1.20 & \small 5.26 & \small 11.65 & \small 4.56 &\small 0.35 & \small 1.29 &\small  1.26\\
        
          \bottomrule
    \end{tabular}   
    }
    \vspace{0.5cm}
    \caption{Automatic and human evaluation performance for pre-trained language models on \dataset. 
    \textbf{B1.}: BLEU1; \textbf{B2.}: BLEU2; \textbf{B3.}: BLEU3; \textbf{B4.}: BLEU4; \textbf{MET.}: METEOR; \textbf{ROU}$_L.$: ROUGE$_L$; 
    \textbf{FLU.}: Fluency (1--5); \textbf{REL.}: Relevance (0--1); \textbf{COM.}: Complexity (1--3); \textbf{INF.}: Informativeness (1--3). } 
    \label{tab:5.1}
    \vspace{-0.5cm}
\end{table*}

\subsection{Comparison with existing datasets} \label{ss:4.2}

We further compare \dataset with three existing QG datasets: SQuAD~\cite{DBLP:conf/emnlp/RajpurkarZLL16}, the most widely-used dataset for answer-aware QG, and LearningQ~\cite{DBLP:conf/icwsm/ChenYHH18} and InquisitiveQG~\cite{DBLP:conf/emnlp/KoCHDL20}, two similar datasets designed for information-seeking QG. 

Table~\ref{tab:compare} shows the comparison on question and document length, question categories, and the leading question words. The question category is based on the level of cognitive skill defined in the Bloom's Taxonomy. We reuse the analytic results of \citet{DBLP:conf/icwsm/ChenYHH18} for SQuAD and LearningQ. For InquisitiveQG, we analyze question categories by manually annotating 100 sampled questions. 

Our findings are as follows. First, questions in \dataset are much longer than in other datasets. The reason is that natural follow-up questions usually contain additional context that is either a conditional clause to limit the scope of the question, or a summarization of the user's understanding of the context.
 Such additional context is often given before the actual questioning sentence to make the whole follow-up question more complete and clear. The include of additional context makes \dataset closer to real-world question-asking. 
Second, \dataset has a more balanced distribution of questions in terms of cognitive skills, and a high percentage of questions (~36\%) in high cognitive levels such as \textit{applying} and \textit{creating}. This makes \dataset significantly different with SQuAD, which is designed to test the reading comprehension ability on low cognitive skill level (\textit{i.e.} \textit{remembering}). 
Although InquisitiveQG also contain a high percentage of high-level questions, the key distinction is that their questions are written by crowd-workers instead of natural-occurring, which results in questions that are typically short and generic (\textit{e.g.}, \textit{Is there a particular example?}). 
LearningQ collect real questions from an online educational platform, therefore containing a large portion of clarification questions. Compared with \dataset, the source contexts of LearningQ (course materials and video captions) are much noisy and considerably longer, making it hard to model and evaluate the problem of FQG. 

\section{Follow-up Question Generation} \label{ss:5}

In this section, we evaluate the ability of three pre-trained language models to generate follow-up questions via fine-tuning, while Section~\ref{ss:6} explores large language models' ability via prompting. Through comprehensive evaluation, we discover the strengths and limitations of current models for follow-up question generation and identify areas ripe for future research.






\subsection{Models}\label{ss:5.1}
We choose three generation models that have shown state-of-the-art results on answer-aware QG: BART~\citep{DBLP:conf/acl/LewisLGGMLSZ20}, T5~\citep{DBLP:journals/jmlr/RaffelSRLNMZLL20}, and GPT-Neo \citep{gpt-neo}. We use \textit{Huggingface} to implement BART-large and T5-base models, and fine-tune these two models on the training set of \dataset by predicting the follow-up question given the concatenation of the initial question and the answer as input\footnote{\texttt{Initial Question <SEP> Answer}}. We use the \textit{aitextgen}\footnote{\url{https://github.com/minimaxir/aitextgen}} library for implementing GPT-Neo, and the input sequence for fine-tuning this model is the concatenation of the initial question, answer and follow-up question\footnote{\texttt{Initial Question <SEP> Answer <QUS> Follow-up Question}} . In the testing time, only initial question and answer is given\footnote{\texttt{Initial Question <SEP> Answer <QUS>}}. 

The batch size for BART, T5 and GPT-Neo is 8, 8 and 16, and we fine-tune for 10 epochs. 
We use Adam \citep{DBLP:journals/corr/KingmaB14} as the optimizer, with a learning rate of 5e-5 for all models. All the models are training on 1 RTX-4080 GPU. Table~\ref{tab:model} shows the details of the models. 
\begin{table}[htbp]
\resizebox{0.48\textwidth}{!}{
    \centering
    \begin{tabular}{cccc}
    \toprule
        Model & Hidden Dimension & Layer & Head  \\
        \hline
        BART-large & 1024  & 24  & 26 \\
        \hline
        T5-base & 768 & 12 & 12 \\
        \hline 
        GPT-Neo &  768 & 12&  12 \\
        \bottomrule
         
    \end{tabular} }
    \vspace{0.2cm}
    \caption{Model details}
    \label{tab:model}
    \vspace{-1cm}
\end{table}

\subsection{Automatic Evaluation}\label{ss:5.2}

We automatic evaluate the generated questions using {BLEU1--4} \citep{DBLP:conf/acl/PapineniRWZ02}, METEOR \citep{DBLP:conf/lavie-agarwal-2007-meteor} and ROUGE-L \citep{DBLP:conf/acl/lin}. 
Results are shown in the top rows of Table \ref{tab:5.1} (Rows 1--3). 
In general, all models achieve much lower scores in automatic metrics, compared with their performance on answer-aware QG. For example, BART achieves a BLEU4 of 21.3 on SQuAD ~\cite{DBLP:conf/acl/PanCXKW20}, while on \dataset{}, it only achieves a BLEU4 of 2.61. Similar observations also hold for T5 and GPT-Neo. This is largely due to the open-ended nature of follow-up question generation. 
Compared with answer-aware QG, where the target answer is given and the questions are mostly factoid, follow-up questions are more open-ended, where the model may generate other plausible questions different from the human references, leading to low performance in $n$-gram based evaluation metrics. This open-ended nature of follow-up questions makes the automatic evaluation less informative. 

\subsection{Human Evaluation}

To better evaluate the quality of generated questions, we conduct human evaluation on 100 randomly sampled pairs in the test set of \dataset. We ask four workers to rate the questions raised by humans and the questions generated by different models for these samples. Workers are blinded by the identity of the models in the annotation. For each question, we ask workers to give ratings on four criteria: \textit{Relevance}, \textit{Fluency} \textit{Complexity}, and \textit{Informativeness}. The detailed criteria are shown in our designed questionnaire in Appendix~\ref{appedix:b}. We average the scores from the workers on each question, reporting averaged performance.

We find that questions generated by BART and T5 achieve comparable scores with human questions in terms of \textit{fluency} and \textit{relevant}. However, the \textit{complexity} and \textit{informativeness} scores are much lower. This indicates that pre-trained models face challenges in solving the key issues of the FQG task, which aims at generating deep and informative questions. Furthermore, the Pearson correlation between automatic and human evaluation results is around 0.38, indicating a weak relationship.  \dataset{} poses a new challenge for developing more faithful question evaluation metrics.

\begin{figure}[!t]
    \centering
    \includegraphics[width=0.45\textwidth]{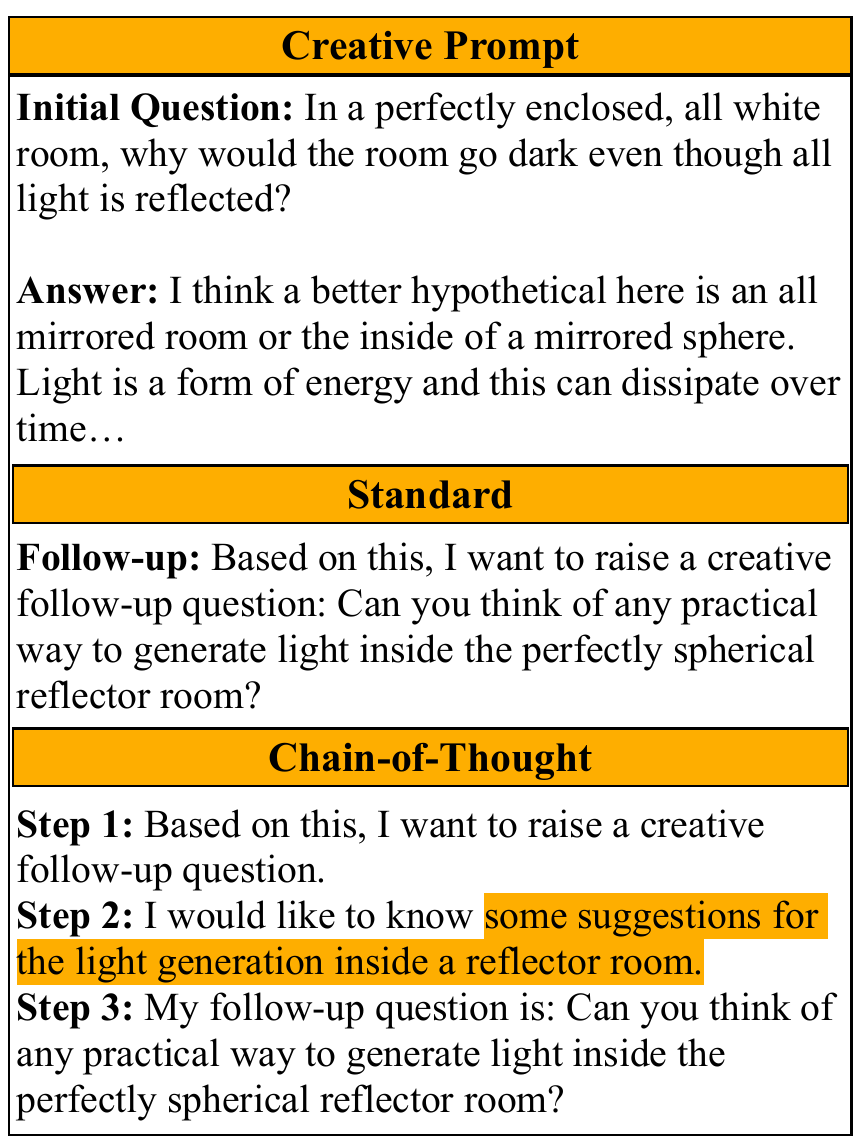}
    \caption{Standard and chain-of-thought \textit{creative} prompt examples. Chain-of-thoughts are highlighted. } 
    \label{fig:stepbystep}

    \vspace{-0.5cm}
    
\end{figure}
\begin{figure*}[!t]
    \centering
    \includegraphics[width=\textwidth]{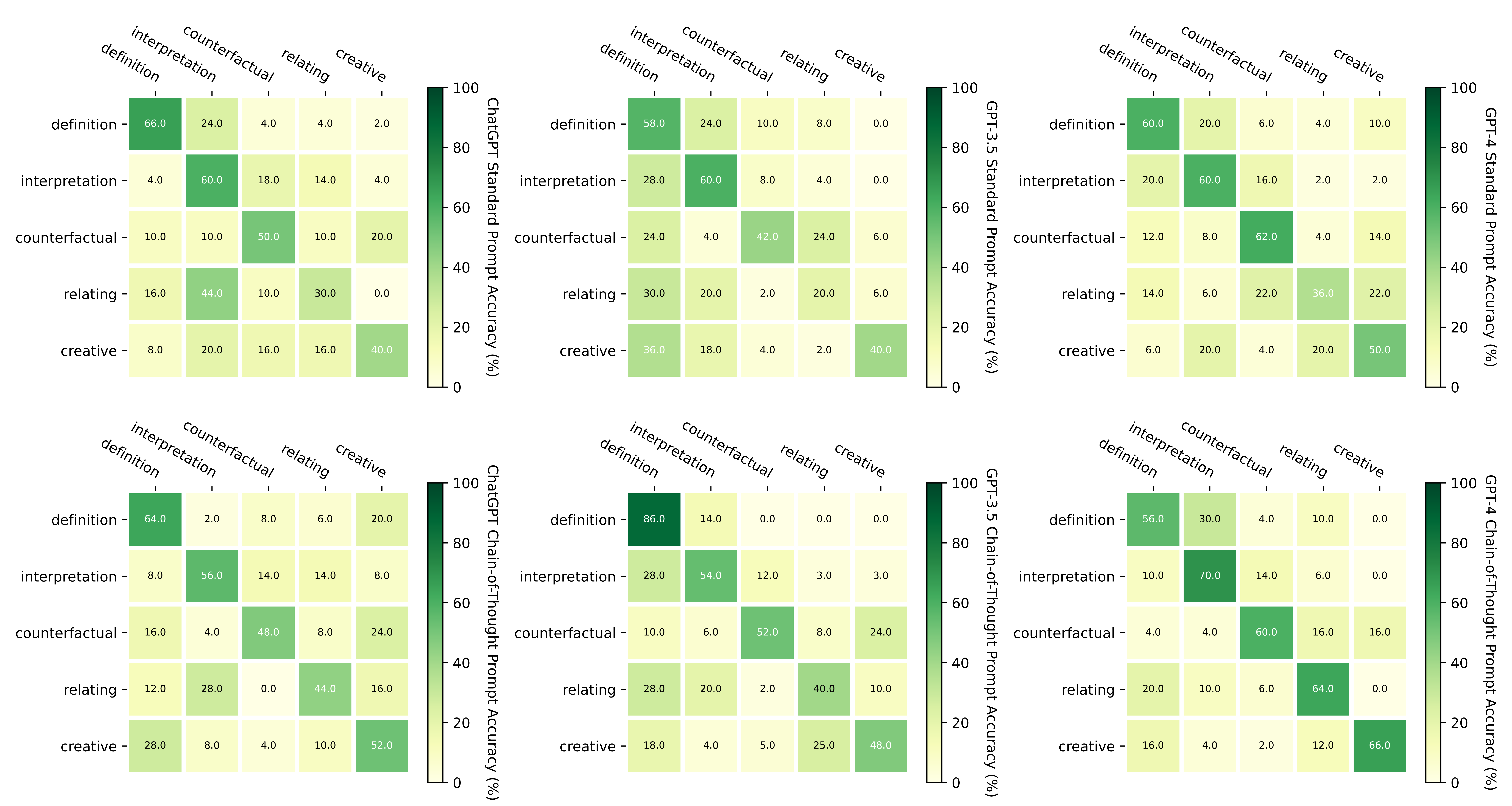}
    \caption{Distribution of generated question types by using different prompts on ChatGPT, GPT-3.5, GPT-4. Vertical labels represent prompts, horizontal labels represent question types.  }
    \label{fig:cot-results}
    
\end{figure*}

\section{Controllable Follow-up QG} \label{ss:6}

We see that the key difficulty in follow-up question generation is due to its open-ended nature. This increases the difficulty of controlling \textit{what to ask} and \textit{how to ask} for  models. We now explore large language models' ability in tackling controllable follow-up question generation via in-context learning. Instead of relying on supervised fine-tuning methods, we adopt the idea of simply ``prompting'' the model with a few input--output exemplars to guide models to generate similar types of follow-up questions. Inspired by ~\cite{saha2023summarization} works on utilizing chain-of-thought reasoning steps for the summarization task, we also give an intermediate reasoning step in a given prompt \textit{(initial question, answer \textit{chain-of-thought}, follow-up question)} to show its effectiveness for question generation.

\subsection{Experimental Setting}
We create standard and chain-of-thought prompts for each type of follow-up question in \S~\ref{ss:5.1}, including \textit{definition}, \textit{interpretation}, \textit{counterfactual}, \textit{relating}, and \textit{creative}. Figure~\ref{fig:stepbystep} illustrates one example of a \textit{creative} prompt for both the standard and chain-of-thought settings\footnote{We list the complete set of exemplars in Appendix~\ref{appendix:c}}. Specifically, chain-of-thought prompts aim to enhance the ability of large language models to accurately control the patterns of follow-up questions during the generation process with an intermediate reasoning step. To evaluate the results of controllability, we elicit language models to generate follow-up questions for 50 sampled (\textit{initial question, answer}) pairs for different types of prompts. This results in 500 generated questions in total. To verify whether prompting an LLM in this way can bring controllability, we manually annotate the question types of the generated questions and calculate the question type accuracy, by comparing whether the types of the generated questions match the  prompt type. 

\begin{table*}[!t]
\centering
\resizebox{0.95\textwidth}{!}{
\begin{tabular}{p{450pt}}
\toprule 

 \textbf{Initial Question: }Why are the size of insects small? Not larger as human beings?  \\
       \textbf{Answer: }In the simplest terms, millions of years ago there were insects that were larger and over 7-8 feet long. It had to do with the higher percentage of oxygen in the atmosphere back then. But as the levels of gases in the atmosphere stabilized and Oxygen decreased, so did the size of these insects... \\
        \textbf{Human: }May I ask, why was there a higher percentage of oxygen? Was it because there were more plants and no human pollution? And why did it go down? \\ \midrule
\end{tabular}
}

\resizebox{0.95\textwidth}{!}{
\begin{tabular}{p{210pt}p{210pt}}
    \textbf{ChatGPT (CoT):} Are there any other examples of animals being affected by changes in atmospheric gases? & \textbf{ChatGPT (Standard):}  Are there any mechanisms that insects have involved to counteract the lack of oxygen in the air?  \\
            \textbf{GPT3.5 (CoT): }What specific factors cause the decrease in oxygen levels that ultimately led to the decrease in insect size? &\textbf{GPT3.5 (Standard):} Why did the oxygen concentration decrease overtime?   \\ 
            \textbf{GPT-4 (CoT): }How did higher oxygen concentrations in the past contribute to the growth and survival of giant insects like {the ones} in prehistoric times? & \textbf{GPT4 (Standard):} Could insects theoretically be engineered to have more effective mechanisms, allowing them to grow larger in our current atmosphere?   \\
            \bottomrule
\end{tabular}
}
\vspace{0.2cm}
\caption{Examples of model-generated follow-up questions via standard and chain-of-thought relating prompts. }

\label{tab:case}
\vspace{-0.5cm}
\end{table*}

\subsection{Result Analysis}


We evaluate chain-of-thought prompting on ChatGPT, GPT-3.5 (text-davinci-003), and GPT-4, respectively. Figure~\ref{fig:cot-results} shows the distribution of generated question types by utilizing standard and chain-of-thought prompts for the large language models. 
First, we observe that generating the same type of questions for the given prompts for low-level question types is relatively simpler for large language models, as compared to higher-order question types. For example, all three models have relatively higher accuracy ($\sim$60\%) in generating \textit{definition} and \textit{interpretation} questions when given corresponding standard or chain-of-thought prompts. However, accuracies drop ($\sim$40\%) when generating \textit{relating} or \textit{creative} questions. 

Secondly, the evaluation of different language models indicates that GPT-4 outperformed other models in terms of controllable question generation tasks, particularly for high-order question generation. GPT-4 achieves around 66\% accuracy in generating \textit{creating} questions while ChatGPT and GPT-3.5 only reaches around 50\% when using chain-of-thought prompts. 

Third, our findings indicate that incorporating a chain-of-thought reasoning step results in an improvement in controllable accuracy when generating follow-up questions, particularly for higher-order question types. Notably, GPT-4 showcases an approximately 16\% increase in accuracy in generating ``creative'' questions when compared to using standard prompts alone. However, ChatGPT and GPT-3.5 still exhibit relatively lower accuracy in controlling high-level questions, even with the utilization of chain-of-thought prompts. These results suggest potential future directions for further advancements in addressing the challenge of controlling and improving accuracy in generating high-level follow-up questions for these models.

\subsection{Case Study}
To give a clear understanding of the differences between follow-up questions raised by humans and large language models, we compare several model-generated questions with different prompts and human-raised questions in our dataset \dataset{}. Table~\ref{tab:case} shows one \textit{relating} human-raised follow-up question and six model-generated questions with \textit{relating} standard and chain-of-thought prompts. More examples are shown in Appendix~\ref{appendix:case} In general, we observe that all models are able to generate fluent and logical follow-up questions based on the given context \textit{(initial question, answer)}. However, they still have some limitations compared with human-written questions. 

First, humans are capable of proposing \textit{relating} questions by providing new and specific relevant examples. For instance, when presented with a context regarding a higher percentage of oxygen, humans can generate additional factors not mentioned in the given context, such as ``increased plant population'' or ``human-induced pollution,'' as shown in Table~\ref{tab:case}. However, both ChatGPT and GPT-4 struggle to perform this task effectively. While GPT-4 may generate associations with new terms, the generated content appears to be more generic and lacks the specific and concrete examples that humans provide.

Second, LLM-generated questions tend to follow a formal and formulaic pattern, which contrasts with the characteristics of our dataset that are layer-person friendly and conversational in nature. This underscores another challenge of using this prompting approach with LLMs to control the style of generated questions. 

\section{Conclusion and Future Work}

We explore the problem of information-seeking follow-up question generation by presenting \dataset, a dataset consisting of 3K \textit{(initial question, answer, follow-up question)} tuples that represent real-life human question-asking, including rich pragmatic functions and diverse cognitive skills. We then propose question generation models on this data via fine-tuning and \textit{chain-of-thought} prompting. Extensive evaluations demonstrate several difficult aspects of follow-up question generation, showing \dataset is a challenging dataset that deserves future investigation. Future works include how to promote higher-order deep questions, how to control the generation process, how to improve the evaluation metrics, and how to adapt follow-up QG in practical applications such as education.  

\section*{Acknowledgement}
This research has been supported by the WING research group of the National University of Singapore. We would like to thank Professor Kan for his support, and anonymous reviewers for their valuable feedback on how to improve the paper.

\section*{Limitations}

We acknowledge several limitations in our work. First, follow-up questions are difficult to evaluate with current automatic evaluation metrics, especially judging whether the questions are seeking new information. Despite human evaluation is involved in this work, it is time-consuming and costly, and also it is difficult to reproduce and guarantee the evaluation consistency. 

Second, since the data is collected from the online question-answering forum, the pragmatic functions we found in \dataset may not cover all types of follow-up questions in real-life. Although \dataset covers diverse types of follow-up questions from low to high cognitive levels, follow-up questions raised in other scenarios (\textit{e.g.}, in the classroom, in paper review, in conversation) might be different and are worthwhile to explore. For example, \textit{criticizing} questions rarely appear in our dataset, probably because in forum QA, the questioners are often less knowledgeable than the answerer in the domain they are asking for. However, in paper reviewing, \textit{criticizing} questions may be more commonly seen. 



Third, for the modeling part, we focus on revealing the limitations of state-of-the-art large language models in follow-up question generation. Although we design one method to improve the generation via \textit{chain-of-thought} prompting, they are quite straightforward and only slightly contribute to generating deeper and more informative questions. More specialized model designs should be explored in the future to improve this, such as modeling the reasoning chain or discourse structure.

\bibliography{anthology,custom}
\bibliographystyle{acl_natbib}

\clearpage
\appendix

\section{Human Annotation Guideline}
\label{sec:appendix}

\subsection{Job Description}

The aim of this data annotation task is to select high quality (Knowledge Goal Question, answer, follow-up question) tuples from online forums. The raw data was crawled from an online educational forum \textit{Reddit} in the topic of \textit{``explainlikeimfive''}, which aims to provide lay person-friendly explanations for real-life questions. We collected many (question, answer) pairs from the forum, together with the replies to the answer. A common type of reply to the answer is \textit{asking follow-up questions}. We selected out those follow-up questions and format each data entry as a triple of (initial question, answer, follow-up question). An example is as follows: \\

\noindent\textbf{Initial Question: }When a disease infects you, most of them aim to kill you. But if they kill you, they die too. So why would they want to kill themselves? \\

\noindent\textbf{Answer: }They don’t aim to kill you, they just survive and spread or not. Some diseases may be new to a particular environment, and the host may be unprepared for a new disease thus resulting in worse sickness and death for the host. Obviously , in general it does not suit the disease if the host dies too quickly and thus does not get a chance to pass it on but near in mind that the disease may have a natural reservoir elsewhere. \\

\noindent\textbf{Follow-Up Question: }If so why harm the host? If the disease could have peace with the body then it could multiply and spread. \\

However, there is much noise in the data we collected. For example, the follow-up question is not a valid question, the data entry contains sensitive information and the follow-up question may be not relevant to the question or the answer. \textbf{Your job is to help us identify and remove those noisy samples. }\\

For each HIT, we will show you one triple of (question, answer, follow-up question), you need to read the question, answer and follow-up question carefully and then answer the following three questions based on your judgement. \\

Q1: Do you think the follow-up question is a valid question? \\

\indent A. Yes B. No \\

Q2: Do you think the (initial question, answer, follow-up question) contains sensitive information?\\

\indent A. Yes B. No\\

Q3: Do you think the follow-up question is related to the original question and the answer? \\

\indent A. Strongly related B. Related C. Slightly related D. Not related\\

When you answered the three questions, click submit to jump to the next sample.

\subsection{Detailed Guidelines}
\subsubsection{Guideline for Q1}
Q1: Do you think the follow-up question is a valid question? \\
A. Yes B. No 
\\

\noindent\textbf{Guideline:} The follow-up question might contain multiple sentences but it should consist of at least one valid question. A valid question must be in \textit{question format} and \textit{ask meaningful information}, including Wh-questions (what/why/where/etc.), open-ended questions, probing questions and etc. Examples for \textit{invalid questions}: ``10000 meters? really?'', which are often used in conversational speech to express feelings instead of asking new information. \\

\noindent\textbf{Examples that are considered INVALID question:}
\\

\noindent\textbf{Initial Question: }Eli5: Why do seahorses still existed?\\

\noindent\textbf{Answer: }That’s not how evolution works. Evolution is survival of the fittest, not survival of the most useful. Seahorses will keep existing as long as they have food and reproduce faster than they are hunted. \\

\noindent\textbf{Follow-up Question: }Ohhh, I can't believe it. Seahorses will keep existing, really? \\

\noindent\textbf{Reason: } ``Seahorses will keep existing, really?" is in question format but is invalid, because it is not asking information but just expressing suspicious feeling. \\

\noindent\textbf{Initial Question: }Why do tapes have screen tearing and splitting, but modern media doesn’t?\\

\noindent\textbf{Answer: }Information on a VHS was an analog signal. To fit all of it on a tape in a linear form would end up with miles of tape. So they record the information in diagonal stripes across the tape, which is much wider than it would need to be if it just held a single track. This is then played back through a spinning head that is place at the same angle relative to the tape as the information on it. The tape moved by the head as the head spins, reading each stripe in a fluid motion, creating the smooth video you see on playback. Pausing it stops the tape moving, but not the spinning head. So the image you see is whatever that spinning head is reading repeated over and over. Hence the "between frames" effect.\\

\noindent\textbf{Follow-up Question: }That’s right, I remember watching a Technology Connections video about how the head itself had to be skewed because of that. I remember as a kid looking inside the VCR and wondering “what is that thing, is it broken?” Now I know.\\

\noindent\textbf{Reason: } ``what is that thing, is it broken?" is just the questioner's memory, not a question asking information in the original question or the answer. \\

\noindent\textbf{Examples that are considered VALID question:}\\

\noindent\textbf{Initial Question: }ELI5: Why is the sea calm in the mornings? \\

\noindent\textbf{Answer: }There are two types of waves which can turn a flat sea into a rougher one - swell waves and wind waves. Swell waves can arrive at any time of day, but because wind waves are generated by the wind, they only develop when the wind begins to blow steadily. Since wind speeds are often low at night, and increase during the daytime, wind waves often die out during the night, leading to a relatively flat sea (perhaps with swell waves) in the early morning. During the day, the wind waves increase in size as the wind speed increases, leading to a rougher, more choppy, sea surface during the afternoon and evening. \\

\noindent\textbf{Follow-up Question: }Now it’s time for my question. Why are winds always weak in the morning and very strong during the day? \\

\noindent\textbf{Reason: }The follow-up question is a ``Why'' question, asking specific reasons about the change of the winds during the day. Therefore, it is a valid question. 

\subsubsection{Guideline for Q2}

Q2: Do you think the (initial question, answer, follow-up question) contains racism, hate speech, sexual topics, offensive or rude comments?\\

\noindent A. Yes B. No \\

\noindent\textbf{Guideline:} Examples for racism comments: ``It's credit to your race'', ``Black people will not understand''. Examples for hate speech: ``He should go back to where he comes from'', `` All Mexicans are rapists''. Examples for offensive or rude comments: ``Women are not suitable for working in IT field.'', ``Gay will never understand''.   \\

\noindent\textbf{Examples that contain INAPPROPRIATE information:} \\

\noindent\textbf{Initial Question: }ELI5: Do Asexual people not feel orgasms or pleasure?\\

\noindent\textbf{Answer: }Asexual people don't feel sexual attraction. That is, the thing that makes a sexual person look at someone else and think "I want to have sex with that person". Asexual people can still feel sexual desire or sexual arousal, but it usually won't be targeted towards someone in particular. I had an asexual roommate, and she did feel sexual arousal, but didn't feel sexual attraction. So, in the rare time that she masturbated, she would be able to become aroused by porn or sexual material, but her arousal wasn't centered in imagining herself performing the actions or having those actions performed on her. \\

\noindent\textbf{Follow-up Question: }Is it like, how? Do their brains work in a way that disallows them of sexual thinking? I don't mean anything bad, believe me, but I can't understand how something as primal and ingrained in our genes as reproducing is nullified by a factor of personality.\\

\noindent\textbf{Reason: }Sexual topic is involved in this data entry, which can be considered as the sensitive data. \\

\subsubsection{Guideline for Q3}

Q3: Do you think the follow-up question is related to the initial question or the answer? \\

\noindent A. Strongly Related B. Related C. Slightly Related D. Not Related \\

\noindent\textbf{ Guideline:} ``Strongly Related'': The follow-up question asks specific definition, particular reasons or meanings in the original question and the answer. The information asked is all included in the original question or the answer. ``Related'': The follow-up question mainly asks information occurred in the original question or the answer but also involves other new information. ``Slightly Related'': The follow-up question mainly asks other cases but the cases are relevant to the original question or the answer. ``Not Related'': The follow-up question asks nothing relevant to the original question or the answer.   \\

\noindent\textbf{Examples that are STRONGLY RELATED to the original question:} \\

\noindent\textbf{Initial Question:} ELI5: How is 2FA security by-passed? \\

\noindent\textbf{Answer:} Various possible methods. If someone inadvertently shared their 2fa setup key (or a situation like having their phone stolen and the authentication app backed up), that could be used alongside the original password to gain access. SIM swapping is another possible tactic assuming the first password and the victims cell phone number is already known: if the attacker calls the carrier and requests a replacement SIM/ESIM delivered to the attacker, the attacker could then take advantage of any voice/text message 2fa prompts. \\

\noindent\textbf{Follow-up Question:} And can anything be done to prevent a SIM swap?\\

\noindent\textbf{Reason:} The follow-up question is asking for elaboration on the solutions about how to prevent a SIM swap, and ``SIM swap'' appears in the answer. In addition, there is no other new information in the follow-up, and thus it can be considered as strongly related. \\

\noindent\textbf{Examples that are RELATED to the given answer.} \\

\noindent\textbf{Initial Question:}ELI5: What is an NFT?\\

\noindent\textbf{Answer:}They’re kind of like digital trading cards. You buy a thing that represents a digital thing (an image or tweet or whatever else), and then you can turn around and sell that token for hopefully more money than you paid for it. Some important things to note: * Owning the token provides you with no practical rights to the thing (for example, you can’t prevent others from using or displaying it) * The token is safe from counterfeiting because it’s backed by the block-chain (no one else can steal your token or produce a fake one) * Because it’s backed by the block-chain, every transaction that happens with your token uses a *lot* of electricity, which is a problem.\\

\noindent\textbf{Follow-up Question: }Is it possible to monetize the usage or display of the NFT? Like royalties for memes?\\

\noindent\textbf{Reason: }This follow-up question mainly asks information about NFT, but related to the case ``royalties for memes''. Therefore, it can be considered as the ``Related'' question. \\

\noindent\textbf{Examples that are SLIGHTLY RELATED to the question or the given answer.}\\

\noindent\textbf{Initial Question: }ELI5: Why turning something off, then on again actually fixes issues?\\

\noindent\textbf{Answer: }Turning something off brings it back to a known good state, and powering it up puts it through a known good set of steps. During the time it was on, errors and junk data and black magic might have occurred, giving bad data to good processes, resulting in poor performance.\\

\noindent\textbf{Follow-up Question: }Is it practically the same for software when uninstalling and then reinstalling it? \\ 

\noindent\textbf{Reason: }This follow-up question is asking whether the information in the answer is also suitable for other cases. The main goal for this question is to ask the principle of ``software install'' but also mentions the information in the answer. Therefore, it is slightly related to the given answer.\\ 

\noindent\textbf{Examples that are NOT RELATED to the given answer.}\\

\noindent\textbf{Initial Question: }ELI5: Why does light get absorbed in black holes if photons are massless?\\

\noindent\textbf{Answer: }Photons are “massless”, but not really. They’re still physical and can be impacted by gravity. This is why light can `curve’ around a dense object like a planet. Black Holes have crazy amount of gravity, and literally traps light inside.\\

\noindent\textbf{Follow-up Question: } Could you also explain the theory on Newton's law?   \\ 

\noindent\textbf{Reason: }Newton's law is nothing related to the question or the answer. Therefore, it can be considered as ``NOT RELATED'' case. \\

\subsection{Quality Check}

We will check the quality of your data at the back-end in the Amazon-Turk. If 80\% of your annotated data entries are invalid, we will probably reject all your annotated data samples. Instead, if your annotated data samples are all in line with our standards, we will give you a bonus as an extra reward. In detail, to control the annotation quality, the test samples will be inserted into each batch. According to the accuracy of the test samples, if the accuracy is lower than 60\%, the annotation from this participant will be rejected.

\section{Human Evaluation Questionnaire} \label{appedix:b}
Table~\ref{tab:6.1.2} shows the human evaluation questionnaire. 
\begin{table}[htbp]
    \centering
    \begin{tabular}{p{200pt}}
    \hline
           \small \textbf{Questionnaire}\\
           \hline
           \small \textbf{Q1.} Whether this follow-up question is relevant to the original question and the answer? \\
           $\circ$ \small Yes $\circ$ \small No\\ 
           \hline
           \small \textbf{Q2.} Whether this follow-up question contains the following errors: \\
          \small  $\square$ correct $\square$ redundant $\square$ invalid $\square$ context copy  $\square$ wrong semantic collocation $\square$ others \\ \hline 
           \small \textbf{Q3.} Whether this follow-up question requires reasoning to generate? \\
            $\circ$ \small Yes, and very hard $\circ$ \small Yes, but slightly simple  $\circ$ \small No\\ \hline
            \small \textbf{Q4.} Whether this follow-up question reveals new information? \\
            $\circ$ \small Yes, and a lot $\circ$ \small Yes, but a few  $\circ$ \small No\\ \hline
        
    \end{tabular}
    \vspace{0.5cm}
    \caption{Questionnaire designed for human evaluation, where $\circ$ and $\square$ indicate single-item and multiple-item selection respectively. }
    \label{tab:6.1.2}
\end{table}

\section{Chain-of-thought Prompt}\label{appendix:c}

\noindent\textbf{Definition Prompt}\\
\textit{Question}: ELI5: How do you take a picture/screenshot of a bluescreen? You can write down the message or take a photo of the screen. \\
\textit{Answer}: If it occurs within a virtual machine, then you can screenshot within the host OS...\\
\textit{Step 1}: Based on this, I want to raise a definition question. \\
\textit{Step 2}: I want the information 'virtual machine' to be further clarified.  \\
\textit{Step 3}: My follow-up question is: Thank you for your explanation. I have a few follow-up questions though: What exactly is a virtual machine? \\
\\
\textit{Question}: ELI5: How does the heart compensate when losing a lot of blood? For example a head injury or something more severe?\\
\textit{Answer}: It beats faster to push what blood you do have to where your body still needs it...\\
\textit{Step 1}: Based on this, I want to raise a definition question.  \\
\textit{Step 2}: I want the information 'put pressure on the wound' to be further clarified.  \\
\textit{Step 3}: My follow-up question is: Doesn't it also regulate pressure by constricting blood vessels to keep the flow at the same pressure?\\
\\
\textit{Question}: ELI5 - What exactly is the voltage measurement across your neuron?\\
\textit{Answer}: Voltage is a measure of potential difference...\\
\textit{Step 1}: Based on this, I want to raise a definition question.  \\
\textit{Step 2}: I want the information 'potential difference' to be further clarified.  \\
\textit{Step 3}: My follow-up question is: So what exactly is the potential difference? is it not the $\delta$V between two points of an electric field?\\
\\
\textit{Question}: ELI5: In Archeology, why do they have to dig to get to stuff? Was the ground plane lower then and higher now, or do things sink over time? Pompeii was literally buried in volcanic ash. \\
\textit{Answer}: Rome is a city built upon a city...\\
\textit{Step 1}: Based on this, I want to raise a definition question.  \\
\textit{Step 2}: I want the information 'Rome' to be further clarified.  \\
\textit{Step 3}: My follow-up question is: Interesting point about Rome. What were the previous cities that existed before Rome?\\
\\
\textit{Question}: ELI5: How does a cell phone determine how much charge is left? \\
\textit{Answer}: My understanding is that batteries output a constant voltage until ...\\
\textit{Step 1}: Based on this, I want to raise a definition question.  \\
\textit{Step 2}: I want the information 'voltage regulator' to be further clarified.  \\
\textit{Step 3}: My follow-up question is: is this through using a voltage regulator that accepts variable input voltage? What is keeping the phone from running on less than 3v, as long as the amperage/resistance change accordingly to keep it in working range?\\
\\
\noindent\textbf{Interpretation Prompt}\\
\textit{Question}: Why do car insurance companies charge you more to retain your full right to sue someone?\\
\textit{Answer}: Because if you delegate your right to sue for pain and suffering to them...\\
\textit{Step 1}: Based on this, I want to raise an interpretation question.  \\
\textit{Step 2}: I want the information 'how to delegate the right and the insurance' to be explained.  \\
\textit{Step 3}: My follow-up question is: When selecting a policy, it gives two options of either limit the right or full right. Why can't we just sue someone individually and cut out the insurance as a middleman? \\
\\
\textit{Question}: Why don't dentists and eye doctors have to go to regular medical school?\\
\textit{Answer}: The simplest explanation is that dentistry and optometrists are more about tools...\\
\textit{Step 1}: Based on this, I want to raise an interpretation question.  \\
\textit{Step 2}: I want the information 'masses curve spacetime' to be explained.  \\
\textit{Step 3}: My follow-up question is: Then there is the flip side to my question as well, why are there not more specialized schools? A podiatrist doesn't need to know the same things an OB does but they both have the same gen ed in medical school.\\
\\
\textit{Question}: How can you convert liabilities to equity? \\
\textit{Answer}: There are two primary ways for a company to raise money: taking on liabilities... \\
\textit{Step 1}: Based on this, I want to raise an interpretation question.  \\
\textit{Step 2}: I want the information 'the relationship between liability and equity' to be explained.  \\
\textit{Step 3}: My follow-up question is: Oh ok I see, makes sense now. So essentially there will be more shareholders and thus more equity, but the liabilities would still have to be paid in the long run. Why is this conversion usually seen as a good thing?\\
\\
\textit{Question}: why are fish sometimes exempt from some vegetarian diets?\\
\textit{Answer}: A 'vegetarian' who eats fish would likely be more aptly called a 'Pescatarian'...\\
\textit{Step 1}: Based on this, I want to raise an interpretation question.  \\
\textit{Step 2}: I want the information 'fish and diets' to be explained.  \\
\textit{Step 3}: My follow-up question is:  Why are fish sometimes the exception, health, morality, or otherwise?\\
\\
\textbf{Counterfactual Prompt}\\
\textit{Question}: Why can't we lower ocean levels manually? \\
\textit{Answer}: You could, but the amount of difference...\\
\textit{Step 1}: Based on this, I want to raise a counterfactual question.  \\
\textit{Step 2}: I want to know the counterfactual cases on 'digging for the fossil fuel'  \\
\textit{Step 3}: My follow-up question is:  What if we were to hypothetically be capable to drill down into the Challenger Deep?  \\
\\
\textit{Question}: ELI5: when we react to noises, why do we react faster when they are louder? \\
\textit{Answer}: The startle reflex is a reaction in the brain that kicks the body into ... \\
\textit{Step 1}: Based on this, I want to raise a counterfactual question.  \\
\textit{Step 2}: I want to know the counterfactual cases on 'the level of noises'.  \\
\textit{Step 3}: My follow-up question is:  What makes the louder noise likely to trigger the reflex if we look at the processes in the body?\\
\\
\textit{Question}: If soap and water clean the body, then what makes towel dirt after a bath? \\
\textit{Answer}: When you dry yourself with a course...  \\
\textit{Step 1}: Based on this, I want to raise a counterfactual question.  \\
\textit{Step 2}: I want to know the counterfactual cases on 'alternative methods to avoid making towel dirt'.  \\
\textit{Step 3}: My follow-up question is:  What if I just wrap a towel around my waist and walk around and air dry? \\
\\
\textit{Question}: How are gorillas and similar animals taught sign language? \\
\textit{Answer}: They aren’t. At least not in the.. \\
\textit{Step 1}: Based on this, I want to raise a counterfactual question.  \\
\textit{Step 2}: I want to know the counterfactual cases on 'language without repeating patterns'.  \\
\textit{Step 3}: My follow-up question is: What is language if not repeating patterns to communicate with another?\\
\\
\textit{Question}: What happens if you get sued for an amount of money significantly higher than you could ever actually pay? \\
\textit{Answer}: It’s why they sell umbrella coverage...\\
\textit{Step 1}: Based on this, I want to raise a counterfactual question.  \\
\textit{Step 2}: I want to know the counterfactual cases on 'umbrella coverage'.  \\
\textit{Step 3}: My follow-up question is:  What if you don't have umbrella coverage?\\
\\
\textbf{Relating Prompt}\\
\textit{Question}: How does a USB-C charger stay in the port?\\
\textit{Answer}: It’s round and smooth on the outer surface... \\
\textit{Step 1}: Based on this, I want to raise a relating question.  \\
\textit{Step 2}: I want to know the relationship between the case of 'usb-c' and to relevant cases 'micro-b'.  \\
\textit{Step 3}: My follow-up question is:  How come usb c is soo much better than micro b? \\

\textit{Question}: What does this actually mean when people say private people tell you so little about themselves but you think you know a lot about them? \\
\textit{Answer}: People tend to fill in the blanks...\\
\textit{Step 1}: Based on this, I want to raise a relating question.  \\
\textit{Step 2}: I want to know the relationship between the case of a 'private person' and relevant cases of the new descriptions of being private.  \\
\textit{Step 3}: My follow-up question is: Is there a more specific word for this beyond being private?   \\
\\
\textit{Question}: How is putting LED screens on grocery store cooler doors, instead of simple glass doors, a profitable choice?\\
\textit{Answer}: The idea is that you save energy...
\textit{Step 1}: Based on this, I want to raise a relating question.  \\
\textit{Step 2}: I want to know the relationship between the case of 'putting LED screens on cooler doors' and the case of 'simple glass doors'.  \\
\textit{Step 3}: My follow-up question is: Except how is that better than a glass door you can see through?     \\
\\
\textit{Question}: How can choosing pictures of certain objects (stop signs, cross walks, cars, etc.) prove that you are not a robot on websites?  \\
\textit{Answer}: Because how bots select images on ...\\
\textit{Step 1}: Based on this, I want to raise a relating question.  \\
\textit{Step 2}: I want to know the relationship between the case of 'software made by the bot' and to new relevant cases of 'human-like bot'. \\
\textit{Step 3}: My follow-up question is: But why haven't the bot makers made their software behave more like a human?    \\
\\
\textit{Question}: Why do some countries' currencies go into million for such a small amount of value?  \\
\textit{Answer}:  For example, in Thailand...
\textit{Step 1}: Based on this, I want to raise a relating question.  \\
\textit{Step 2}: I want to know the relationship between the case of 'domination' and new relevant cases.\\
\textit{Step 3}: My follow-up question is:  But why they don't do any re-domination to make things easier?    \\
\\
\textbf{Creative Prompt}\\
\textit{Question}: In a perfectly enclosed, all-white room, why would the room go dark even though all light is reflected? \\
\textit{Answer}: I think a better hypothetical ...\\
\textit{Step 1}: Based on this, I want to raise a creative question.  \\
\textit{Step 2}: I want to know new solutions or suggestions for the case of 'light generation inside a reflector room'. \\
\textit{Step 3}: My follow-up question is:  Can you think of any practical way to generate light inside the perfectly spherical reflector room?   \\
\\
\textit{Question}: During a live televised sports match, how do they get the replay footage edited in so quickly?  \\
\textit{Answer}: I design control rooms and their replays systems...\\
\textit{Step 1}: Based on this, I want to raise a creative question.  \\
\textit{Step 2}: I want to know new solutions or suggestions for the 'career path'.\\
\textit{Step 3}: My follow-up question is: What career path led you to that?    \\
\\
\textit{Question}: How are PGP signatures used to verify users on forums?  \\
\textit{Answer}: You have to have access to the private ...\\
\textit{Step 1}: Based on this, I want to raise a creative question.  \\
\textit{Step 2}: I want to know new solutions or suggestions for the 'private key'.\\
\textit{Step 3}: My follow-up question is:  How would they know you had the private key?   \\
\\
\textit{Question}: How does your brain 'realize' that you're home?  \\
\textit{Answer}: I kinda feel the same..\\
\textit{Step 1}: Based on this, I want to raise a creative question.  \\
\textit{Step 2}: I want to know new solutions or suggestions for the 'unexplained human instinct'.\\
\textit{Step 3}: My follow-up question is: why would you say that it can't be explained?  \\
\\
\textit{Question}: How do Small Business Owners Pay Themselves?  \\
\textit{Answer}: You pay yourself as an employee and so you are taxed ... \\
\textit{Step 1}: Based on this, I want to raise a creative question.  \\
\textit{Step 2}: I want to know new solutions or suggestions for 'how to define a phenomenon'.\\
\textit{Step 3}: My follow-up question is: Any idea how they define “reasonable”?

\section{Case study}\label{appendix:case}

Examples of human-written follow-up questions and model-generated questions for standard and chain-of-thought prompts for definition (Table~\ref{tab:case1}), interpretation (Table~\ref{tab:case2}) counterfactual (Table~\ref{tab:case3}) and creative (Table~\ref{tab:case4}) types. 

\begin{table*}[!t]
\centering
\resizebox{0.9\textwidth}{!}{
\begin{tabular}{p{450pt}}
\toprule 

 \textbf{Initial Question: } Why are PS4 and PS5 unable to read PS1 or PS2 discs?  \\
       \textbf{Answer: } PS1 and some PS2 disks are based on the CD standard. The drive in a PS4 (and I assume PS5) doesn't read CDs. Among other things, it's only equipped with lasers for DVD (650nm) and Blu-ray (450nm). \\
        
        \textbf{Human: }A lot of PS2 games were on DVD though. Why can't the console read those? \\ \midrule

\end{tabular}}
\resizebox{0.9\textwidth}{!}{
\begin{tabular}{p{210pt}p{210pt}}

    \textbf{ChatGPT (CoT):} What exactly is the CD standard? Is it the disc size and shape, or is it the way data is stored on the disc? & \textbf{ChatGPT (Standard):}  What is the difference between CDs, DVDs, and Blu-ray discs in terms of their laser technology and storage capacity?  \\
            \textbf{GPT3.5 (CoT): } What is the difference between the CD standard and the DVD/Blu-ray standards that make the PS4 unable to read the PS1 or PS2 discs? &\textbf{GPT3.5 (Standard):} So does that mean that if the PS4 had the right lasers, it could read the PS1 and PS2 discs?   \\ 
            \textbf{GPT-4 (CoT): }Could you elaborate more on the CD standard? Also, are there any technical reasons why they didn't include the ability to read CDs on the PS4 and PS5 other than cost-saving measures? & \textbf{GPT4 (Standard):} So the reason for the inability to read PS1 or PS2 discs on PS4 and PS5 is due to the different disc formats and laser technology used, correct? Are there no possible software solutions to overcome this hardware limitation or enable backward compatibility? \\
            \bottomrule
\end{tabular}}

\vspace{0.2cm}
\caption{Examples of human written follow-up questions and model-generated questions via standard and chain-of-thought definition prompts. }
\vspace{-0.3cm}
\label{tab:case1}

\end{table*}

\begin{table*}[!t]
\centering
\resizebox{0.9\textwidth}{!}{
\begin{tabular}{p{450pt}}
\toprule 

 \textbf{Initial Question: }What are scripts in terms of code? \\
       \textbf{Answer: }Typically a script refers to a small application written in a high-level programming language that isn't compiled into an executable file which typically automates a process rather than being an application that a user interacts with. \\
        
        \textbf{Human: }So like, say I coded a quick grade calculator in Python. That would be a script right? While an actual game in C++ that I could run would be a program? \\ \midrule

\end{tabular}}
\resizebox{0.9\textwidth}{!}{
\begin{tabular}{p{210pt}p{210pt}}

    \textbf{ChatGPT (CoT):} Can you give an example of a script and how it is used for automation? & \textbf{ChatGPT (Standard):}  Can you give some examples of common tasks that are automated by scripts in various fields like IT, finance, or web development?  \\
            \textbf{GPT3.5 (CoT): }So when you compile a script, does that mean it is converted into an executable file and can then be used as an application? What are the other benefits of compiling a script? &\textbf{GPT3.5 (Standard):} How do you compile a script?   \\ 
            \textbf{GPT-4 (CoT): }Can you give some examples of scripting languages and their common uses? & \textbf{GPT4 (Standard):} What are some examples of scripting languages and what kinds of processes can be automated with scripts? \\
            \bottomrule
\end{tabular}}

\vspace{0.2cm}
\caption{Examples of human written follow-up questions and model-generated questions via standard and chain-of-thought interpretation prompts. }
\vspace{-0.3cm}
\label{tab:case3}

\end{table*}

\begin{table*}[!t]
\centering
\resizebox{0.9\textwidth}{!}{
\begin{tabular}{p{450pt}}
\toprule 

 \textbf{Initial Question: }What difference do the vocal cords of a person who can sing have in comparison to someone who can't? \\
       \textbf{Answer: }Absolutely nothing. The vocal cords are an instrument, the same as a trumpet or clarinet. Anyone is capable of singing well, the difference between someone who is good and who is bad is practice, technique, and for some, proper training. There have been studies proving that anyone can be taught to sing despite the fact that they have any practice in singing or not.\\
        
        \textbf{Human: } Are there any biological differences in someone's singing ability when it comes to how well they can hear the true pitch that they are putting out? Not vocal chord-related obviously, but the idea that anyone can kind of baffles me. I mean my friend. Asking for a friend. \\ \midrule

\end{tabular}}
\resizebox{0.9\textwidth}{!}{
\begin{tabular}{p{210pt}p{210pt}}

    \textbf{ChatGPT (CoT):} What about the influence of genetics on singing ability? Are there any studies or evidence that suggest certain genetic traits may make it easier or harder for individuals to develop singing skills? & \textbf{ChatGPT (Standard):}  If everyone can be taught to sing well, then why do some people seem to have a natural talent for singing?  \\
            \textbf{GPT3.5 (CoT): }If someone has great technique and practice but still has difficulty singing at certain pitches, is it possible that some people just have a lower ceiling of what they can do and others don't? &\textbf{GPT3.5 (Standard):}  If everyone can learn techniques to sing well, then why do some people seem to be talented at singing?  \\ 
            \textbf{GPT-4 (CoT): } Are there any genetic factors that influence people's singing ability? & \textbf{GPT4 (Standard):} If there are no physical differences in vocal cords between good and bad singers, are there any other factors, such as genetics or anatomy, that contribute to someone's natural singing ability or vocal range?  \\
            \bottomrule
\end{tabular}}

\vspace{0.2cm}
\caption{Examples of human written follow-up questions and model-generated questions via standard and chain-of-thought counter-factual prompts. }
\vspace{-0.3cm}
\label{tab:case4}

\end{table*}

\begin{table*}[!t]
\centering
\resizebox{0.9\textwidth}{!}{
\begin{tabular}{p{450pt}}
\toprule 

 \textbf{Initial Question: }Why do we have an internal monologue and how does it work?  \\
       \textbf{Answer: }From what I've read it's believed that some people have a connection between their frontal lobe where conscious thought is processed and their auditory processing center where speech is processed and constructed. Not everyone has this link however. About 25\% of people don't and have no internal monologue. About 25\% of people conduct this internal monologue all the time, and the rest only do part of the time.  \\
        
        \textbf{Human: }This is such a strange concept to me. How do you think without words? My inner monologue never stops. \\ \midrule

\end{tabular}}
\resizebox{0.9\textwidth}{!}{
\begin{tabular}{p{210pt}p{210pt}}

    \textbf{ChatGPT (CoT):}  Could the absence of an internal monologue be beneficial in any way? & \textbf{ChatGPT (Standard):} Could you give me some examples of potential benefits of internal monologue?    \\
            \textbf{GPT3.5 (CoT): }What are the potential benefits of having an internal monologue? &\textbf{GPT3.5 (Standard):} Are there any specific benefits to having an internal monologue?  \\ 
            \textbf{GPT-4 (CoT): }Since there's a considerable difference in the presence of an internal monologue among people, are there any known advantages or disadvantages associated with having or not having an internal monologue? & \textbf{GPT4 (Standard):} Is there a correlation between having an internal monologue and certain personality traits or cognitive abilities, such as creativity or problem-solving skills?   \\
            \bottomrule
\end{tabular}}

\vspace{0.2cm}
\caption{Examples of human written follow-up questions and model-generated questions via standard and chain-of-thought creative prompts. }
\vspace{-0.3cm}
\label{tab:case2}

\end{table*}

\end{document}